%% file: main.tex
\documentclass{article}

\usepackage[preprint]{corl_2026} 

\usepackage{graphicx}
\usepackage{booktabs, multirow}
\usepackage{array}
\usepackage{caption}
\DeclareCaptionFont{capsize}{\fontsize{8}{9}\selectfont}
\usepackage{enumitem}
\usepackage{xcolor}
\usepackage{etoolbox}
\usepackage{amsmath,amsfonts,bm}
\usepackage{amssymb}
\usepackage{soul}
\usepackage{wrapfig}

\newcommand{\fullname}{Gated Memory Policy}
\newcommand{\shortname}{GMP}
\newcommand{\benchname}{MemMimic}
\newcommand{\supp}{Supplementary Materials}

\newcommand{\blue}[1]{\textcolor[rgb]{0, 0.4, 1}{#1}}
\newcommand{\darkblue}[1]{\textcolor[rgb]{0.08,0.37,0.66}{#1}}
\newcommand{\gray}[1]{\textcolor[rgb]{0.4, 0.4, 0.4}{#1}}
\newcommand{\purple}[1]{\textcolor[rgb]{0.47, 0.22, 0.54}{#1}}

\newcommand{\mypara}{\vspace{1mm}\noindent\textbf}

\title{Gated Memory Policy: \\ In-Context Memorization and Adaptation}

\author{
  Yihuai Gao \quad Jeff Jinyun Liu \quad Shuang Li \quad Shuran Song \\
  \vspace{1mm}
  Stanford University \\
  \vspace{1mm}
  \href{https://gated-memory-policy.github.io}{https://gated-memory-policy.github.io}
}

\begin{document}
\maketitle

\vspace{-8mm}

\begin{abstract}

Robotic manipulation tasks exhibit varying memory requirements, ranging from Markovian tasks that require no memory to non-Markovian tasks that demand \emph{in-context memorization} of historical information within a single trial or \emph{in-context adaptation} based on the outcomes of multiple past trials. Surprisingly, simply extending observation histories of a visuomotor policy often leads to a significant performance drop due to distribution shift and overfitting.
To address these issues, we propose Gated Memory Policy (GMP), a visuomotor policy that learns both when to recall memory and what to recall, enabling both in-context memorization and adaptation.
To learn when to recall memory, GMP employs a learned memory gate mechanism that selectively activates history context only when necessary, improving robustness and reactivity.
To learn what to recall efficiently, GMP introduces a lightweight cross-attention module that constructs effective latent memory representations. To further enhance robustness, GMP injects diffusion noise into historical actions, mitigating sensitivity to noisy or inaccurate histories during both training and inference.
GMP achieves a 30.1\% average success rate improvement over long-history baselines on our proposed non-Markovian benchmark MemMimic, a 26.6\% improvement over VLA baselines on MIKASA-Robo benchmark, while maintaining competitive performance on Markovian tasks in RoboMimic. All code, data and in-the-wild deployment instructions are available on our \href{https://gated-memory-policy.github.io}{project website}.
\end{abstract}

\keywords{Memory, In-Context Learning, Imitation Learning}


\input{text/1_introduction.tex}
\input{text/2_related_work.tex}

\input{text/3_method.tex}

\input{text/4_experiments.tex}
\input{text/5_final_sections.tex}


\bibliography{references}


\clearpage
\input{text/supp.tex}

\end{document}

%% file: text/1_introduction.tex
\section{Introduction}
\input{figures/teaser.tex}

Robotic manipulation tasks exhibit a wide range of memory requirements. Simple pick-and-place tasks are largely Markovian \cite{robomimic2021,liu2023libero} and require minimal history. More complex tasks demand \textbf{in-context memorization} through in-trial memory, where the policy needs to recall past actions and visual observations within a single execution episode \cite{cherepanov2025mikasa,chung2025liberomem}. The most challenging tasks require \textbf{in-context adaptation} through cross-trial memory\footnote{\label{footnote:mem-types}In-trial and cross-trial memory are often referred to as working and reference memory in psychology \cite{refmem,frick1995age}.}, which aggregates information across multiple interactions to infer unobservable physical properties of objects and environments \cite{chi2022irp}.
For example, in Fig.~\ref{fig:teaser} (c), the robot needs to infer the object surface friction from repeated casting iterations and adapt its subsequent actions in context.

Designing a policy formulation that handles all memory requirements is challenging. Naively extending observation histories enables non-Markovian behavior but often degrades performance on Markovian tasks.
A longer history substantially increases the input dimensionality, raising the risk of overfitting \cite{torne2025ldp}. This also exacerbates distribution shift between training and deployment. In particular, the presence of noisy or unfamiliar observations in a few history frames can shift the entire input sequence out of distribution.
Additionally, attending to long histories (especially cross-trial memory) incurs significant computational overhead. Together, these issues make history-based policy designs both unreliable and inefficient.

To address these challenges, an effective memory-enabled policy must answer two questions: 

\textbf{When should memory be recalled?} Memory is unnecessary during most of the task execution, and conditioning on it may distract the policy from the current observation, slowing down inference and reducing reactivity. A robust policy should therefore ignore history in these cases.

\textbf{What information should be recalled?} For complex manipulation tasks, it is difficult to specify explicit rules for which information to store and retrieve. For instance, the casting task described above cannot be reduced to either visual or action history alone; instead, it requires inferring object properties by combining visual and action history from multiple interactions.

In this work, we introduce \textbf{\fullname~(\shortname)}, a practical and effective framework toward memory-focused imitation learning: 

To learn \ul{\textit{When}} to recall memory, \shortname~employs a learned \emph{memory gate} that adaptively regulates information flow between historical context and current observations. We calibrate the memory gate by comparing action prediction errors with and without memory on a validation set. The gated policy is able to selectively recall memory, which prevents over-reliance on past observations, improves generalization, and reduces unnecessary computation.

To learn \ul{\textit{What}} to recall efficiently, we leverage a cross-attention module that learns to focus on the most relevant history without additional supervision. Additionally, compared to feeding history to a bi-directional attention network, \shortname~eliminates redundant computation within the history itself, reducing both memory usage and runtime cost.


To rigorously evaluate long-term memory in manipulation, we introduce \textbf{\benchname}, a suite of simulation and real-world benchmarks that explicitly require both in-trial and cross-trial memory. We additionally evaluate \shortname~on RoboMimic, which consists of Markovian tasks with minimal memory requirements. Overall, \shortname~consistently achieves strong performance compared to alternatives, demonstrating a \textbf{30.1\%} improvement on non-Markovian tasks while maintaining competitive performance on Markovian benchmarks.

In summary, our contributions are:  
1) \fullname~(\shortname), an efficient and practical memory-focused policy framework that selectively attends to history on demand, maintaining both robustness and reactivity. \shortname~enables both in-context memorization (in-trial memory) and in-context adaptation (cross-trial memory) without sacrificing performance on Markovian tasks.
2) \benchname, a benchmark designed to evaluate history usage in visuomotor policies with in-the-wild data and a focus on cross-trial memory tasks that are largely overlooked by existing benchmarks.  
3)  A comprehensive empirical study in simulation and the real world, examining memory-related design choices and their impact on performance, robustness, and computational efficiency.

%% file: figures/teaser.tex
\begin{wrapfigure}{r}{0.5\linewidth}
    \centering
    \vspace{-16mm}
    \includegraphics[width=\linewidth]{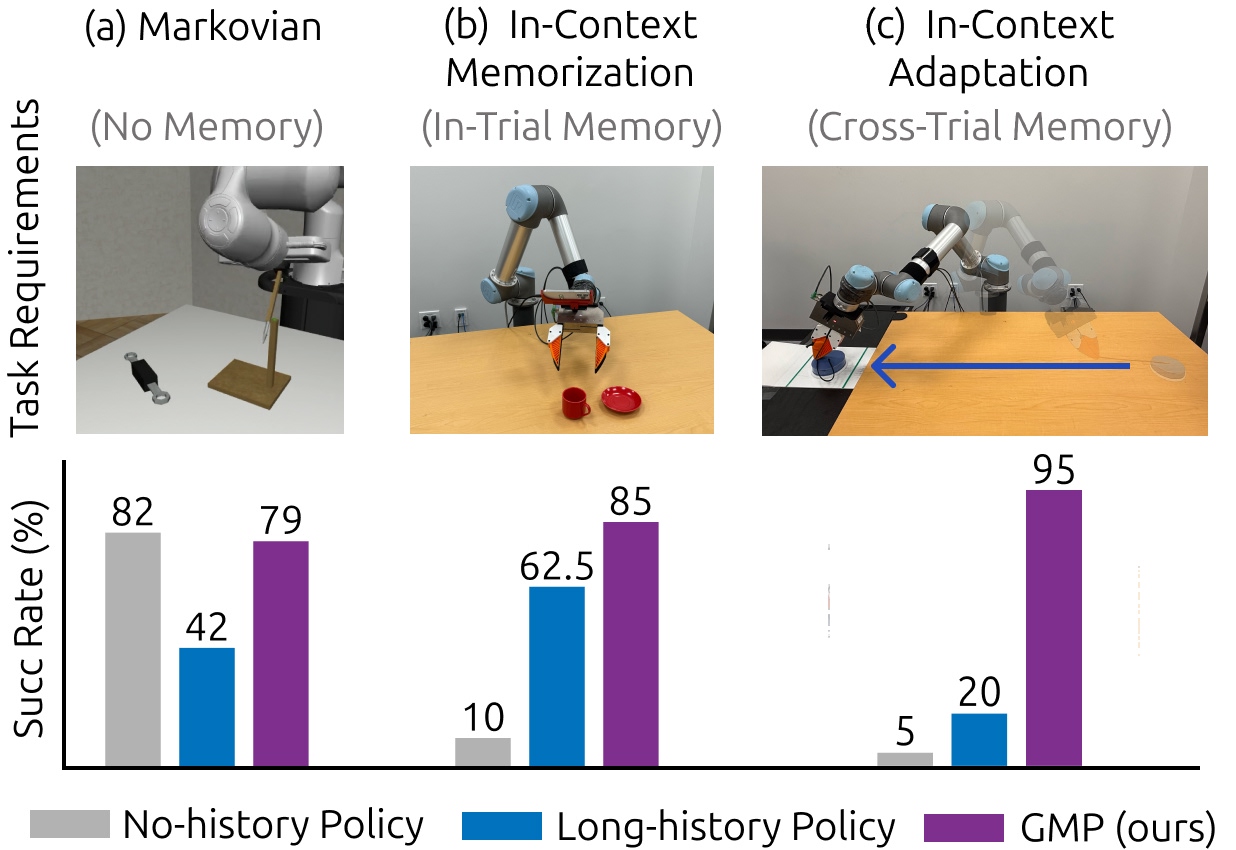}
    
    \caption{\textbf{Memory Requirements} in robotics range from (a) \textit{Markovian} tasks requiring no memory; (b) \textit{in-trial} memory for in-context memorization within a single execution; to (c) \textit{cross-trial} memory that summarizes information across multiple attempts for in-context adaptation\textsuperscript{\ref{footnote:mem-types}}. Naively increasing policy history often degrades policy performance in Markovian tasks and is computationally expensive for cross-trial memory tasks. Our Gated Memory Policy (GMP) selectively recalls task-relevant history, achieving strong performance across tasks.      
    } 
    \label{fig:teaser}
    \vspace{-5mm}
\end{wrapfigure}

%% file: text/2_related_work.tex
\section{Related Work}


\textbf{Structured Memory for Robot Policy} Prior memory-based policies have typically utilized \textit{action parameterization or trajectory tracking}, such as storing keyframe heatmaps~\cite{fang2025sam2act}, referencing object trajectories~\cite{chi2022irp, vandermerwe2025icip}, or visual trace overlays~\cite{zheng2024tracevla}. Alternatively, \textit{semantic and latent representations} can be used to store history, such as environment dynamics estimation~\cite{kumar2021rma}, LLM-generated textual plans~\cite{liu2023reflect}, or vector database retrieval of observation and prompt embeddings~\cite{anwar2024remembr}.
Despite their success, structured memory often requires manual design, such as predefined keypoint tracking, specific keyframe selection, or language planning syntax. This makes it inherently task-dependent and time-consuming to deploy.
In contrast, our proposed method, \shortname, learns memory recall in a task-agnostic space -- raw image observations and robot action trajectories -- ensuring seamless generalizability to new tasks.

\textbf{Generalizable Long Context Visuomotor Policies}
The role of temporal context in visuomotor policies remains a subject of active research. While RNN-based policies have been explored \cite{robomimic2021,fang2019smt,cherepanov2025mikasa}, recent studies indicate that shorter or even no history generally achieves better performance on {Markovian tasks} \cite{chi2024diffusionpolicy,pi0,liu2024rdt,kim24openvla,kim2026cosmos}. However, for {non-Markovian tasks}, including in-context learning \cite{fu2024icrt} and adaptation \cite{liulocoformer,wagenmaker2025behavioralexploration}, long-context policies are essential.

Recent attempts to scale context length leverage efficient memory retrieval and auxiliary training objectives. Some approaches utilize external memory banks to retrieve action and image histories \cite{shi2025memoryvla, lin2025echovla, torne2026mem}, while others employ keyframe-selection \cite{sridhar2025memer} and attention aggregation \cite{hu2025pam} to integrate critical information into the policy input. To enhance policy robustness, noise injection is applied to history tokens during training \cite{chen2025diffusion, song2025history}. Furthermore, prediction of history actions \cite{torne2025ldp} and future videos \cite{li2026causalworldmodelingrobot, li2025unified} is also used as auxiliary tasks to enhance temporal understanding.
However, these methods often need to trade off different capabilities. For example, including memory in high-level VLM planners~\cite{shi2025memoryvla,sridhar2025memer} often lacks fine-grained action-observation alignment, and is therefore unable to infer complex object dynamics and iteratively refine actions. Furthermore, including additional generation objectives slows down policy inference \cite{torne2025ldp,song2025history,li2026causalworldmodelingrobot}, impairing policy reactivity.
Our work addresses these gaps by attending to both action and image tokens within a simple cross-attention module, achieving linear complexity in history length.

%% file: text/3_method.tex
\section{Method}
\subsection{Background: Transformer-based Diffusion Policy}
\label{sec:background}
We adopt the Transformer-based Diffusion Policy~\cite{chi2024diffusionpolicy,liu2024rdt} as our network backbone. Given the current image observation $I_t$ and robot proprioception $P_t$, the policy predicts a trajectory of future actions $A_{t:t+h}=\{A_t, A_{t+1}, \cdots, A_{t+h-1}\}$, where $h$ is the action prediction horizon.
During training, Gaussian noise $\epsilon \sim \mathcal{N}(0,1)$ is progressively added to the ground-truth action $A^0_{t:t+h}$.
We train a denoising network $\varphi_\theta$ to predict the ground-truth action and calculate the action prediction loss as
\begin{equation}
\mathcal{L}_{\text{action}}
    = \mathbb{E}_{A^0_{t:t+h},\epsilon,k}
    \left[
    \|A^0_{t:t+h} - \varphi_{\theta}(A_{t:t+h}^k, I_t, P_t, k)\|_2^2
    \right],
\end{equation}
where $k$ is the diffusion step. At test time, actions are denoised for $K$ steps using the DDIM~\cite{song2020denoising} noise scheduler.

\input{figures/method_architecture.tex}

While Diffusion Policy performs well on Markovian tasks, its performance degrades when history-dependent capabilities are required.
A straightforward way to incorporate longer history is to extend the observation-action window. However, this approach introduces two significant drawbacks:  
\textbf{(1) Overfitting.} As history length $H$ increases, the input space expands while the training data remains fixed, making it harder for the model to generalize. Consequently, the model overfits more easily and performs poorly on long-history sequences at inference time.
\textbf{(2) High computational cost.} In Transformer-based architectures, self-attention over $H$ historical tokens scales quadratically as $\mathcal{O}(H^2)$, making both training and inference increasingly expensive as $H$ grows.

\subsection{Gated Memory Policy}
\label{sec:memory_gate}
We use an MLP to encode a set of $n$ historical action trajectories $\{A_{t-nh:t-(n-1)h}, \cdots, A_{t-h:t}\}$, where each chunk consists of a sequence of $h$ actions. Since encoding images requires significantly more computation and high-frequency frames are often redundant, we sample only one image observation per trajectory, yielding $I_{t-nh:h:t} = \{I_{t-nh}, I_{t-(n-1)h}, \cdots, I_{t-h}\}$. We use the pretrained ViT encoder \cite{dosovitskiy2021vit,tschannen2025siglip2} to extract visual features from each image, followed by multi-head attention pooling (MAP) to aggregate all image patches into a single token, substantially reducing the number of visual tokens.

\textbf{Cross-Attention with Cached Tokens for History Conditioning.}  
Directly concatenating all historical action and visual tokens into a single long sequence and processing them with a Transformer is computationally expensive. 
Inspired by the KV cache in causal Transformers \cite{vaswani2017attention,ott2019fairseq}, we cache a sliding window of $n$ history trajectories, where each trajectory consists of one aggregated image feature and $h$ action tokens, as shown in Fig.~\ref{fig:network} (b). \shortname~can therefore retrieve history tokens without additional computation.



\textbf{Diffusion Noise Augmentation to Reduce Overfitting.}
Directly using clean history for training often leads to overfitting and poor generalization at test time \cite{chen2025diffusion}, due to the expansion of the input space. To mitigate this, we propose to use augmented history actions, obtained by injecting diffusion-scheduled noise into past actions, as a history condition. As shown in Fig.~\ref{fig:network}~(b), at diffusion step $k$, the model predicts future actions $A_{t:t+h}^{k}$ at noise level $k$, conditioned on history actions $A_{t-nh:t}^{k-1}$ at noise level $k-1$, one step cleaner than the future actions. This augmentation exposes the model to a significantly larger and more diverse input space than clean actions alone, improving robustness and generalization.

\textbf{Memory Gating for On-Demand Memory Recall.}
In many manipulation tasks, a long history is informative only at occasional but critical moments. Always injecting history can introduce noise and encourage over-reliance on context that is often unnecessary. We therefore learn a \emph{memory gate} that adaptively modulates the contribution of history at each deployment step.

Given the current image observation $I_t$ and robot proprioception $P_t$, we use an MLP $\phi$ with a sigmoid activation $\sigma$ to produce a binary gate value $\mu_t = \mathbf{1}\{\sigma\left(\phi(I_t, P_t)\right) > 0.5\} \in \{0,1\}.$
As shown in Fig.~\ref{fig:network}(b), the gate $\mu_t$ is applied to the history cross-attention output $\mathbf{h}_{t:t+h}$ before adding it to the residual connection $\mathbf{z}_{t:t+h} \in \mathbb{R}^{h \times d}$,
where $h$ is the prediction horizon and $d$ is the DiT hidden dimension. Finally, $\bar{\mathbf{z}}_{t:t+h}=\mu_t \mathbf{h}_{t:t+h} + \mathbf{z}_{t:t+h}$ is passed to the feed-forward layer.

\input{figures/method_training.tex}

\textbf{Memory Gate Calibration.}
The memory gate is challenging to train in an end-to-end manner: the policy tends to use as much history as possible and overfit to the training data.
Although we can apply a regularization term on $\mu$ to discourage unnecessary history use, it is difficult to find a weight that works well across all tasks. A large weight suppresses history usage and hurts performance on non-Markovian tasks, while a small weight fails to provide sufficient regularization (see Fig.~\ref{fig:ablation_gate_regularization}).

We therefore adopt a self-supervised calibration process to generate labels for the memory gate, as illustrated in Fig.~\ref{fig:training}. We train two policies on the \textbf{training set} $\mathcal{D}_\mathrm{train}$, one with the memory gate always off ($\pi$) and one with it always on ($\pi_\mathrm{mem}$). On the \textbf{validation set} $\mathcal{D}_\mathrm{val}$, we label a timestep as ``memory-required'' if adding history information significantly reduces action prediction error. We then train the memory gate to predict the memory requirement. We finally freeze its weights and retrain the policy on the full dataset. This ensures that the policy uses history information only when necessary during both training and testing.




%% file: figures/method_architecture.tex
\begin{wrapfigure}{r}{0.5\linewidth}
    \centering
    \vspace{-15mm}
    \centering
    \includegraphics[width=\linewidth]{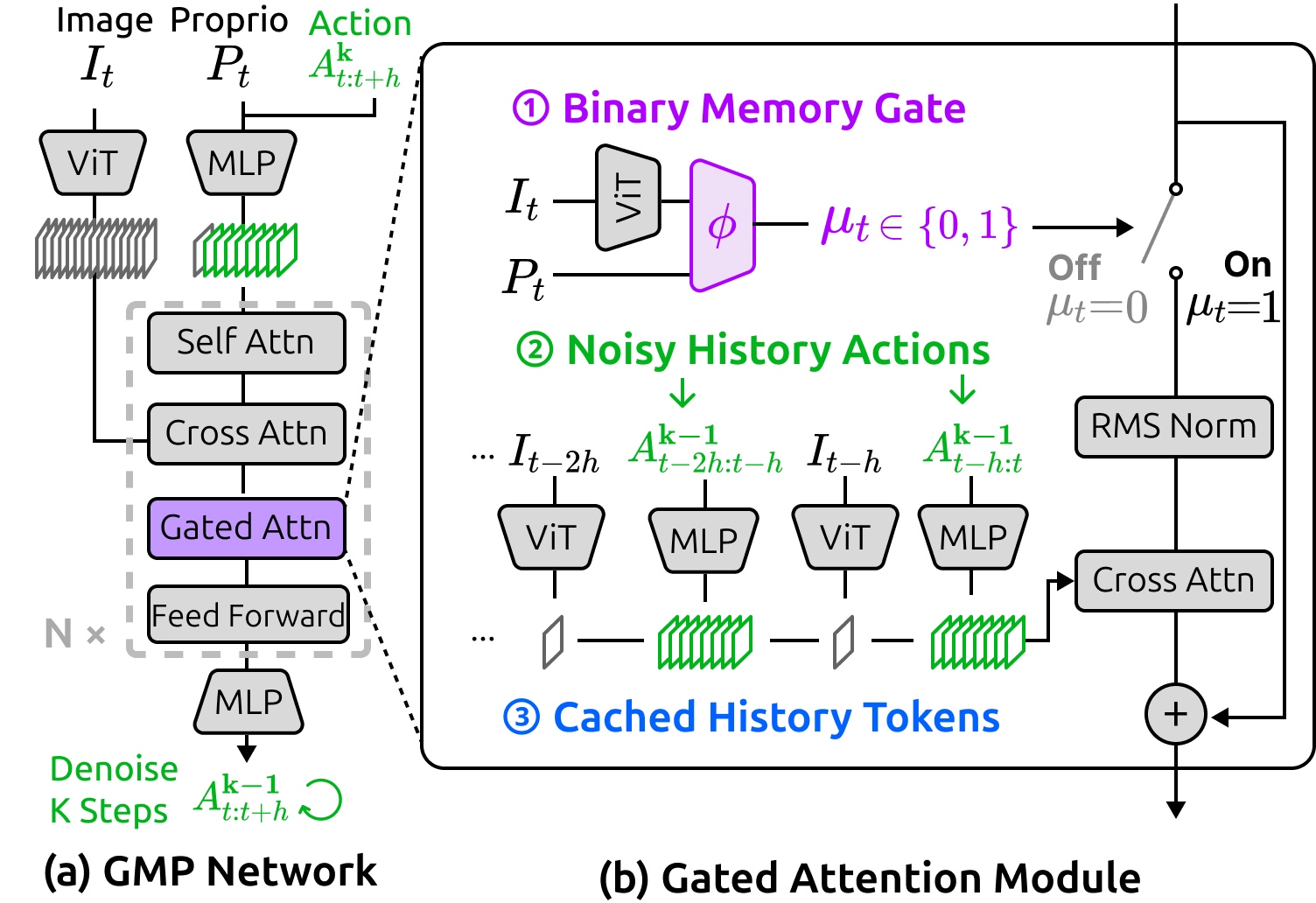}
    \vspace{-10pt}
    \caption{\textbf{\fullname~Network} \textbf{(a)} Based on Diffusion Transformer (DiT) \cite{Peebles2022DiT,liu2024rdt}, we add a gated attention module to selectively recall memory. 
    \textbf{(b)} The gated attention module features three key designs: (1) Binary memory gate $\mu_t$ that determines whether history cross-attention is skipped or applied. (2) Noised history action condition to improve robustness and reduce overfitting. (3) Cached history tokens during inference time to reduce computational cost.}

    \label{fig:network}
    \vspace{-5mm}
\end{wrapfigure}

%

%% file: figures/method_training.tex
\begin{figure*}
    \centering
    \vspace{-8mm}
    \centering
\begin{center}
\includegraphics[width=\linewidth]{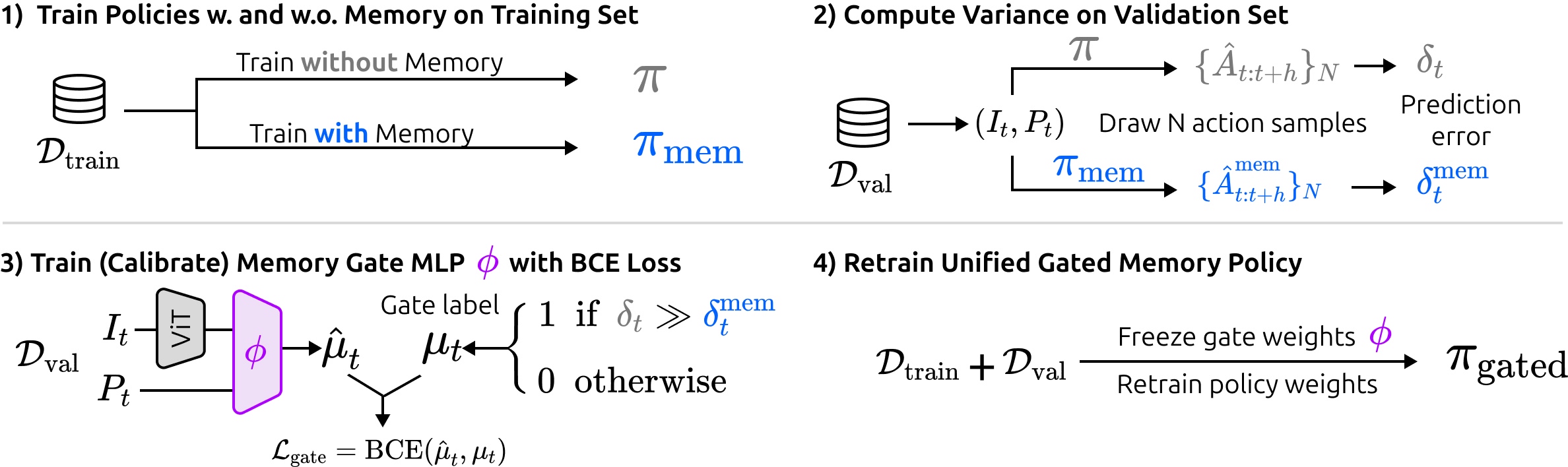}
\end{center}
\vspace{-8pt}
\caption{
\textbf{Memory Gate Calibration.} We split the dataset into $\mathcal{D}_\mathrm{train}$ and $\mathcal{D}_\mathrm{val}$.
\textbf{1)} Train two policies on $\mathcal{D}_\mathrm{train}$ with memory gate always off $\gray{\pi}$ or always on $\blue{\pi_\mathrm{mem}}$.
\textbf{2)} Evaluate the two policies on $\mathcal{D}_\mathrm{val}$ for $N$ rounds and calculate the error of the predicted actions at each timestep $t$.
\textbf{3)} At timestep $t$, if the error of the no-memory policy $\gray{\delta_t}$ is significantly larger than the error of the memory policy $\blue{\delta^\mathrm{mem}_t}$, we label \emph{memory-required} ($\mu_t=1$) for this timestep; otherwise, $\mu_t=0$. This stage is referred to as \textbf{Calibration} of the memory gate. 
\textbf{4)} We freeze the memory gate weights and retrain the policy on the full dataset to obtain the final Gated Memory Policy $\pi_\mathrm{gated}$.  }
\label{fig:training}
\vspace{-5mm}
\end{figure*}

%% file: text/4_experiments.tex

\input{figures/exp1_match_color.tex}

\input{figures/exp2_discrete_place_back.tex}

\section{Experiments}
\mypara{Benchmarks.} 
Here we briefly summarize the benchmark tasks; details of each task can be found in the supplementary material.

\underline{In-trial and Cross-trial Memory:} We propose \benchname, a series of simulation and real-world non-Markovian tasks to evaluate a policy's in-context memorization (in-trial memory) and in-context adaptation (cross-trial memory) capabilities. In-trial memory tasks are shown in Fig.~\ref{fig:experiment1_match_color}, \ref{fig:experiment2_discrete_place_back}, \ref{fig:experiment3p_in_the_wild_flip_and_place_back}, and cross-trial memory tasks are shown in Fig.~\ref{fig:experiment4_iterative_pushing}, \ref{fig:experiment5_iterative_flinging}, \ref{fig:experiment6_iterative_casting}.

\underline{Markovian Tasks:} 
We evaluate the proposed method on RoboMimic~\cite{robomimic2021}, which does not require history information, verifying that memory mechanisms do not harm performance on standard manipulation tasks. Results are shown in Fig.~\ref{fig:exp7_robomimic}.

\underline{Additional In-trial Memory Benchmark:}
We evaluate our method on \textbf{MIKASA-Robo~\cite{cherepanov2025mikasa}}, a memory-intensive benchmark to compare with VLA baselines~\cite{cherepanov2025mikasa,shi2025memoryvla,kim24openvla,octo_2023,qu2025spatialvla}, shown in Fig.~\ref{fig:exp10_mikasa}.

\input{figures/exp3p_in_the_wild_flip_and_place_back.tex}

\mypara{Baselines.} We compare \shortname~with six baselines, including Diffusion Policy (DP)~\cite{chi2024diffusionpolicy}, Past Token Prediction (PTP)~\cite{torne2025ldp}, and BC-RNN~\cite{robomimic2021} with various history lengths. For a fair comparison, all the models in our experiments use SigLIP2-B/16~\cite{tschannen2025siglip2} with pre-trained weights as the image encoder. Please refer to the \supp~for details.

\input{figures/exp4_iterative_pushing.tex}

\input{figures/exp5_iterative_flinging.tex}

\input{figures/exp6_iterative_casting.tex}

\input{figures/exp7_robomimic.tex}

\input{figures/exp10_mikasa.tex}


\mypara{Finding 1: Naive history extension degrades performance.}
We observe that simply increasing the history length of robot policies degrades performance on Markovian tasks. As shown in Fig.~\ref{fig:exp7_robomimic}, [Long-hist DP] (120 history steps) performs significantly worse than [Mid-hist DP] (16 history steps) and [No-hist DP]. This degradation persists even for methods explicitly designed to leverage long histories, such as [Long-hist PTP], which still struggles to effectively filter irrelevant or redundant history information to overcome the overfitting issue.

For non-Markovian tasks, while naive long-history policies perform well on simple tasks such as \texttt{Match Color} and \texttt{Discrete Place Back} (Fig.~\ref{fig:experiment1_match_color}, \ref{fig:experiment2_discrete_place_back}), they are insufficient for more complex real-world tasks (Fig.~\ref{fig:experiment3p_in_the_wild_flip_and_place_back}, \ref{fig:experiment6_iterative_casting}) or cross-trial memory tasks (Fig.~\ref{fig:experiment4_iterative_pushing}, \ref{fig:experiment5_iterative_flinging}, \ref{fig:experiment6_iterative_casting}).
In contrast, our method selectively attends to task-relevant history and introduces diffusion-noise-augmented history actions during training, making it robust to complex environments.

\mypara{Finding 2: History cross-attention selectively attends to the right modality at the right time.}
For the in-trial memory task \texttt{Match Color} (Fig.~\ref{fig:experiment1_match_color}), at timestep $t=80$, when the robot places the cube back, the highest attention is assigned to timestep $t=48$, when the colors were first observed. This shows that the model automatically identifies and attends to key frames without explicit supervision.
For cross-trial memory tasks such as \texttt{Iterative Pushing} (Fig.~\ref{fig:experiment4_iterative_pushing}), attention in Trial 4 focuses on the outcomes of Trial 2 and Trial 3, corresponding to overshoot and undershoot behaviors, respectively. By attending to these past outcomes, the policy adapts its actions in context to achieve successful pushes in subsequent trials.

\mypara{Finding 3: Memory gate helps skip memory recall when unnecessary for Markovian and non-Markovian tasks.}
When calibrating the memory gate (Fig.~\ref{fig:training}) for Markovian tasks, the action prediction error $\blue{\delta^\mathrm{mem}_t}$ of the policy with memory $\blue{\pi_{\mathrm{mem}}}$ is similar to $\gray{\delta_t}$ of the policy without memory $\gray{\pi}$; thus, the memory gate is almost always off ($\mu=0$) and the policy ignores history information, leading to competitive performance against the no-history policy. Results on the RoboMimic \cite{robomimic2021} benchmark are shown in Fig.~\ref{fig:exp7_robomimic}.

Moreover, even for tasks that require history information, the memory gate is off ($\mu=0$) most of the time (for example, 73\% in \texttt{Match Color} and 58\% in \texttt{Iterative Pushing}) and opens only when necessary, further speeding up policy inference. See Finding 5 for speed comparisons. 

\mypara{Finding 4: Calibration on validation set is crucial for memory gate training.} An alternative approach to training a binary gate together with the policy model is to use straight-through estimators (STE) \cite{ste,vqvae}. Specifically, this involves using the binarized value in the forward pass while treating it as a continuous value in backpropagation. During training, since overfitting to more history generally reduces training error, the memory gate value tends to grow toward 1. This undermines policy generalizability and robustness on Markovian tasks.
\input{figures/ablation_gate_regularization.tex}
To mitigate this, one approach is to add a regularization loss to the memory gate value to suppress overuse. However, the gate value tends to drop to 0 if the regularization weight is large, impairing the performance of non-Markovian tasks, as shown in Fig.~\ref{fig:ablation_gate_regularization}.

In contrast, calibrating the memory gate independently on a validation set avoids the dilemma mentioned above. By comparing the action prediction error with and without memory on validation samples, we can identify the critical moments that truly necessitate historical information, instead of overusing or underusing it.


\input{figures/ablation_runtime}
\mypara{Finding 5: \shortname~maintains low computational cost by leveraging the cross-attention and gating mechanisms. }
\label{finding5:computational_cost}
We run policy inference for 100 rounds on an NVIDIA RTX 3080 GPU and report the average inference time with standard deviation, as shown in Fig.~\ref{fig:inference_time_comparison}. The computational cost of the self-attention DiT baseline scales quadratically with the context length, thus taking significantly longer as the history length increases. To reduce the inference time, \shortname~employs two techniques: 1) moving all the history tokens into a separate cross-attention module, making the computational cost scale linearly with the history length (see \darkblue{[GMP (Gate On)]}); 2) skipping the history attention when the memory gate is off, reducing inference time to a short constant when memory is not needed (see \purple{[GMP (Gate Off)]}). 

\input{figures/ablation_noise}
\mypara{Finding 6: The added diffusion noise in history actions improves policy robustness.} 
Our proposed noise injection strategy \purple{[Diffusion Noising]} uses the one-step-cleaner history actions $A_{t-nh:t}^{k-1}$ at diffusion step $k$ both during training and testing. We compare our method with the following baselines: Using [No Noise] in both training and testing makes the policy overly reliant on clean history actions, and thus not robust to distribution shifts during evaluation; adding [Random Level] noise in both training and testing prevents the policy from accessing clean history actions and might result in losing critical information; [Diffusion Forcing] \cite{chen2025diffusion} injects random noise levels during training but no noise during testing, leading to inconsistency between training and testing. The results on \texttt{Iterative Pushing} are shown in Fig.~\ref{fig:ablation_noise}. 

%% file: figures/exp1_match_color.tex
\begin{figure*}
    \begin{center}
    \includegraphics[width=\textwidth]{figures/jpg/exp1_match_color.jpg}
    \end{center}
    \vspace{-10pt}
    \caption{\textbf{Task 1: Match Color (Sim)}. 
    \textbf{(a)} The robot picks up a cube while observing four randomly colored bins. After lifting the cube, the bin colors are randomized; the robot must place the cube into the bin that matches the cube's original color. 
    \textbf{(b) Baselines:} Diffusion Policy (DP) \cite{chi2024diffusionpolicy} and Past-Token-Prediction (PTP) \cite{torne2025ldp} are evaluated across no, medium, and long-history [nh, mh, lh] settings. Our method, \shortname, uses the [lh] configuration.
    \textbf{(c) Visualization:} We visualize the gated cross-attention weights for each 8-step action chunk. At $t=80$, when the robot is placing the cube back, the attention focuses primarily on $t=48$, when the robot first observed the bin colors. Blue indicates the memory gate is on ($\mu_t=1$); gray indicates the gate is off ($\mu_t=0$).
    }
    \vspace{-4mm}
    \label{fig:experiment1_match_color}
\end{figure*}

%% file: figures/exp2_discrete_place_back.tex
\begin{figure*}
    \begin{center}
    \includegraphics[width=\textwidth]{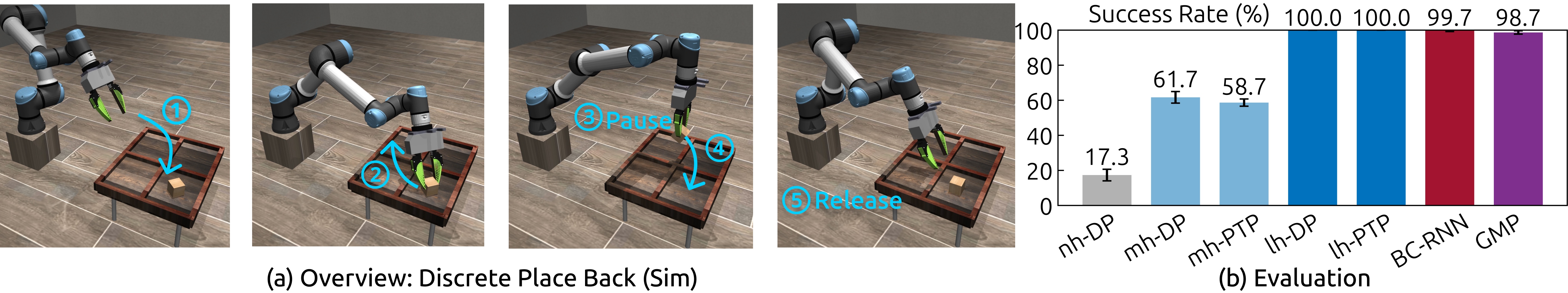}
    \end{center}
    \vspace{-3mm}
    \caption{\textbf{Task 2: Discrete Place Back (Sim)}. 
    \textbf{(a)} The cube is randomly placed in one of the 4 bins. The robot must pick up the cube, hold it in the air for 2 seconds (creating a memory challenge), and return it to the original bin.
    }
    \vspace{-5mm}
    \label{fig:experiment2_discrete_place_back}
\end{figure*}

%% file: figures/exp3p_in_the_wild_flip_and_place_back.tex
\begin{figure*}
    \vspace{-4mm}
    \centering
\begin{center}
    \includegraphics[width=\textwidth]{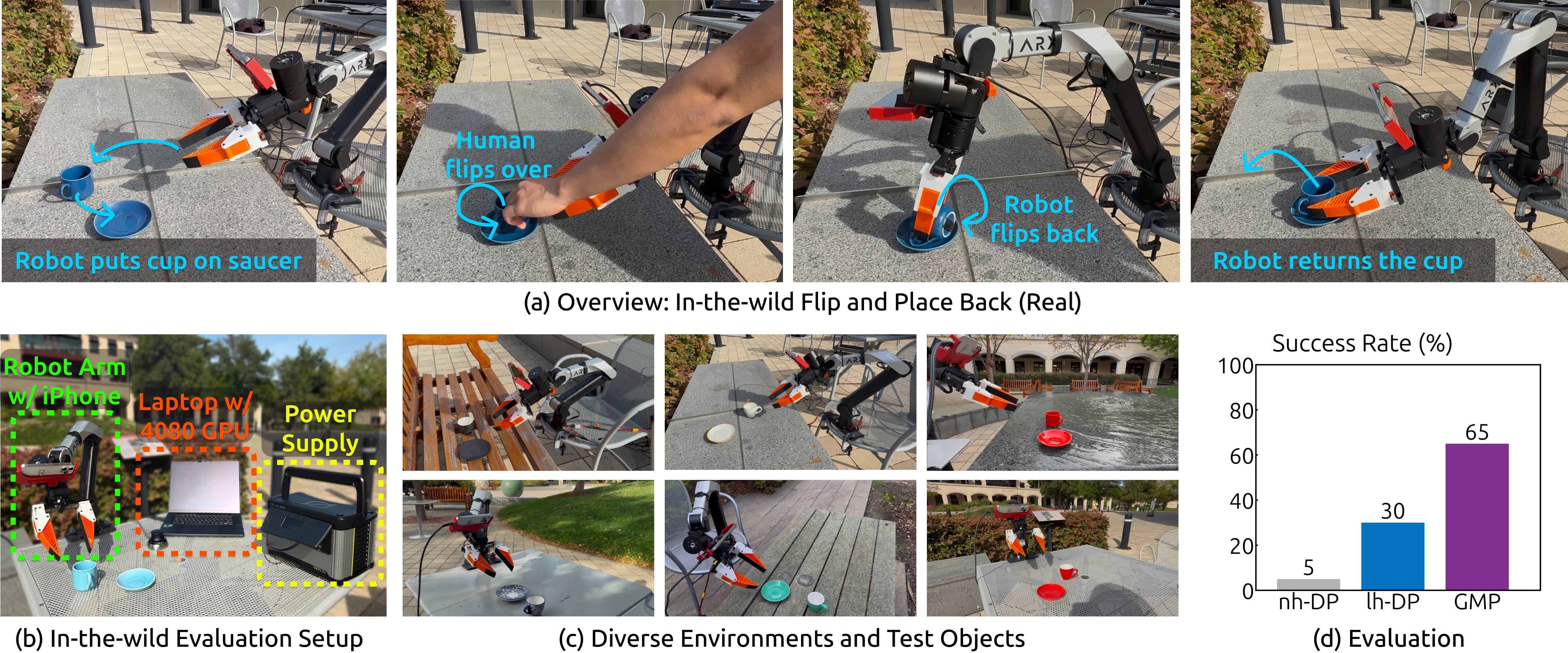}
    \end{center}
    \vspace{-3mm}
    \caption{\textbf{Task 3: Flip and Place Back (Real)}. 
    \textbf{(a)} After the robot picks up the cup and places it on the saucer, a human flips it over by 90 degrees. The robot must flip it back then return it to the original position.
    \textbf{(b)} We use an portable ARX X5 robot arm with an iPhone for visual observation. 
    \textbf{(c)} We evaluate our policy on 6 challenging unseen environments and over 10 cups, showing robustness and generalizability across objects and scenes.
    }
    \vspace{-5.5mm}
    \label{fig:experiment3p_in_the_wild_flip_and_place_back}
\end{figure*}

%% file: figures/exp4_iterative_pushing.tex
\begin{figure*}
    \begin{center}
    \includegraphics[width=\textwidth]{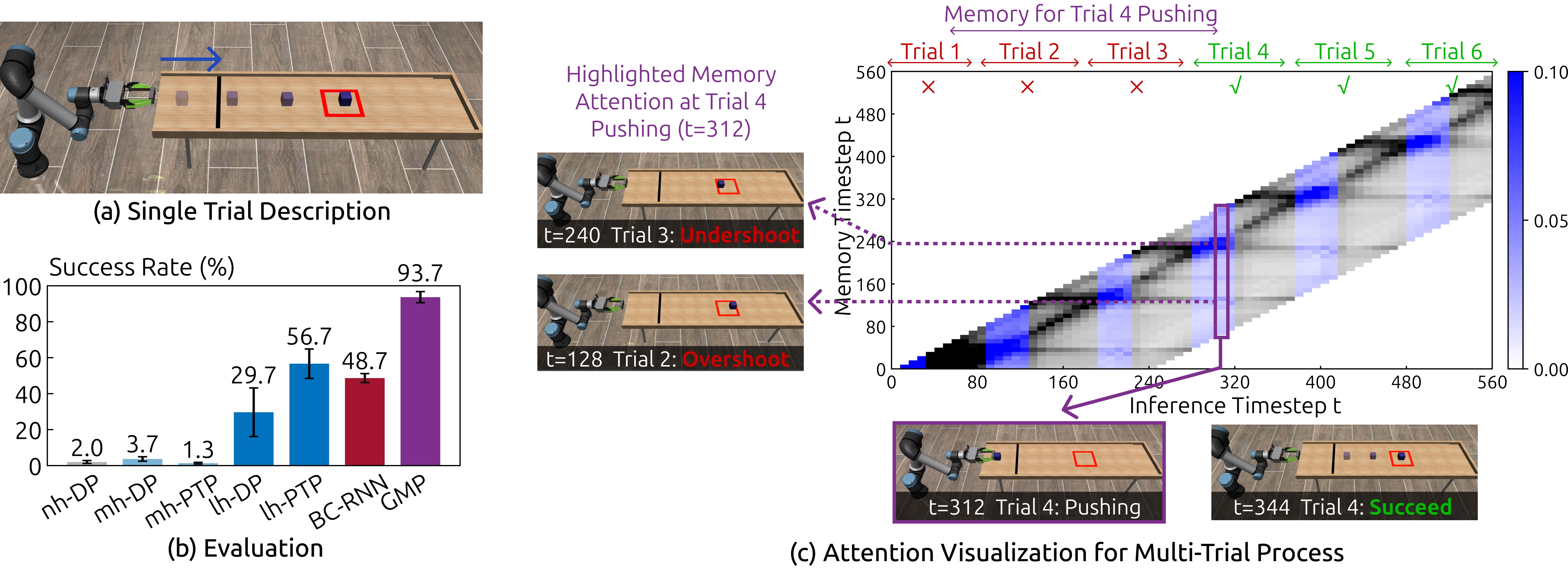}
    \end{center}
    \vspace{-10pt}
    \caption{
    \textbf{Task 4: Iterative Pushing (Sim)}. 
    \textbf{(a)} The robot pushes a cube (with unknown friction) into the red box over 6 trials per episode. The environment resets after each trial. By observing how far the cube moves, the policy learns the physical dynamics and adjusts the pushing velocity accordingly for subsequent trials.  
    \textbf{(c)} Visualization of gated cross-attention weights for each 8-step action trajectory throughout one episode. Blue indicates the memory gate is on ($\mu_t=1$) and gray indicates the memory gate is off ($\mu_t=0$). 
    During the pushing stage in Trial 4, attention focuses primarily on the outcomes from Trial 3 (undershoot) and Trial 2 (overshoot). The policy uses these past outcomes to adjust its action and achieve a successful push. 
    }
    \vspace{-1mm}
    \label{fig:experiment4_iterative_pushing}
\end{figure*}

%% file: figures/exp5_iterative_flinging.tex
\begin{figure*}
    \vspace{-4mm}
    \begin{center}
    \includegraphics[width=\textwidth]{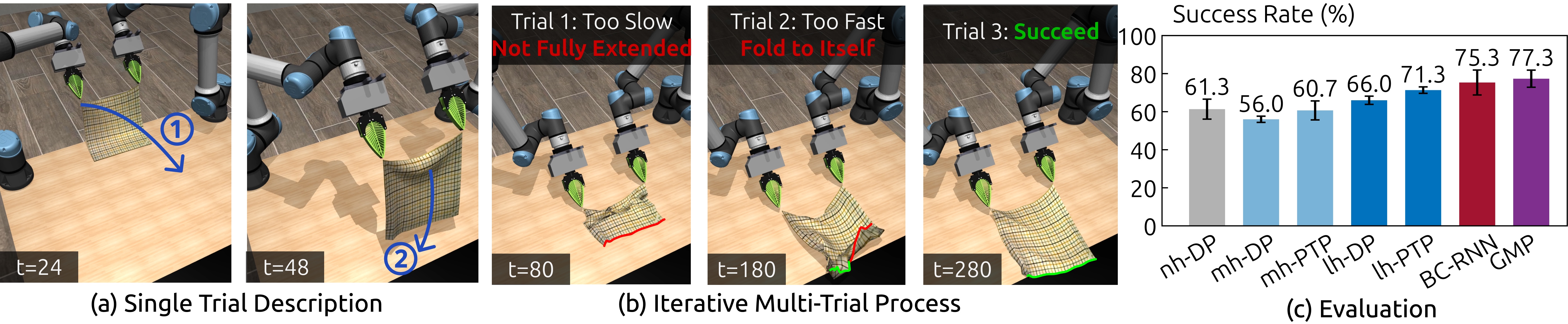}
    \end{center}
    \vspace{-4mm}
    \caption{
    \textbf{Task 5: Iterative Flinging (Sim)}. 
    \textbf{(a)} The robot flings a cloth (with unknown mass) so that the far edge lands on the black target. 
    \textbf{(b)} Multi-attempt flinging. Flinging too slowly leaves the cloth not fully extended; flinging too fast causes the cloth to fold back on itself. The policy learns to adjust the flinging velocity to match the unknown mass of the cloth.
    }
    \vspace{-5mm}
    \label{fig:experiment5_iterative_flinging}
\end{figure*}

%% file: figures/exp6_iterative_casting.tex
\begin{figure*}
    \begin{center}
    \includegraphics[width=\textwidth]{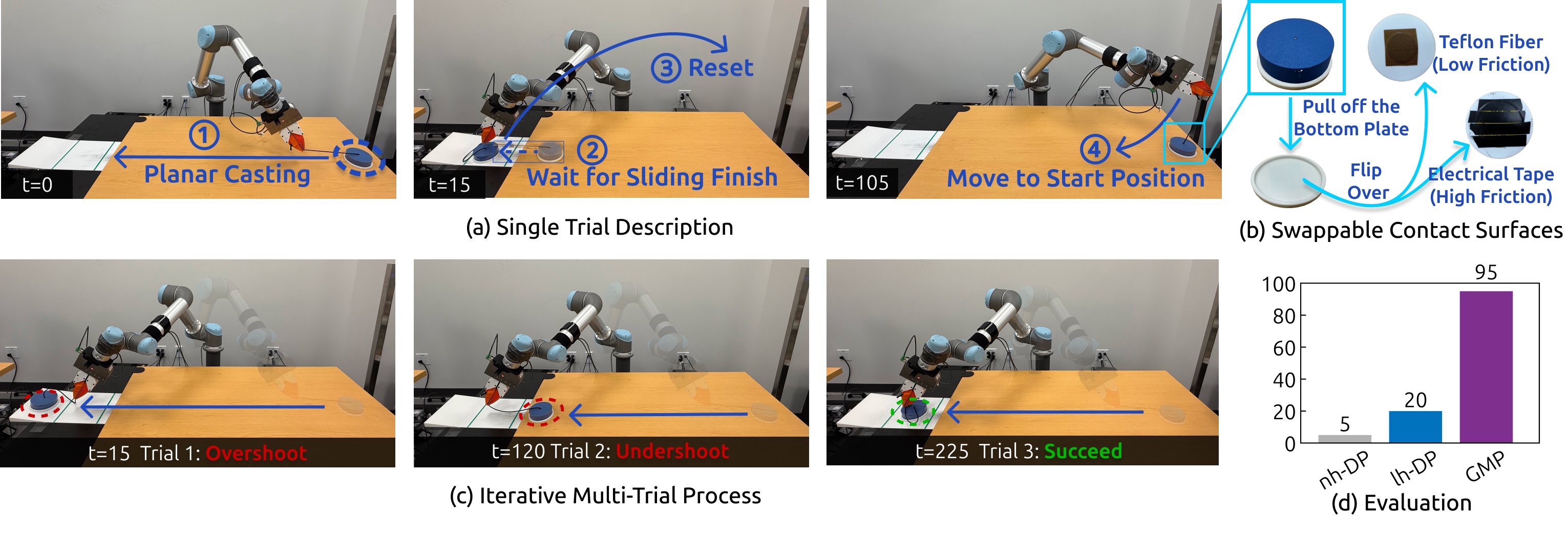}
    \end{center}
    \vspace{-6mm}
    \caption{
    \textbf{Task 6: Iterative Casting (Real)}. 
    \textbf{(a)} The robot casts the object (with unknown friction) so that it stops between the two lines. After the object stops, the robot moves back for the next trial. 
    \textbf{(b)} Two object contact surfaces with different friction coefficients are used across episodes, requiring the policy to infer object dynamics and apply different casting velocities. }
    \vspace{-5mm}
    
    \label{fig:experiment6_iterative_casting}
\end{figure*}

%% file: figures/exp7_robomimic.tex
\begin{figure*}
    \begin{center}
    \includegraphics[width=0.95\textwidth]{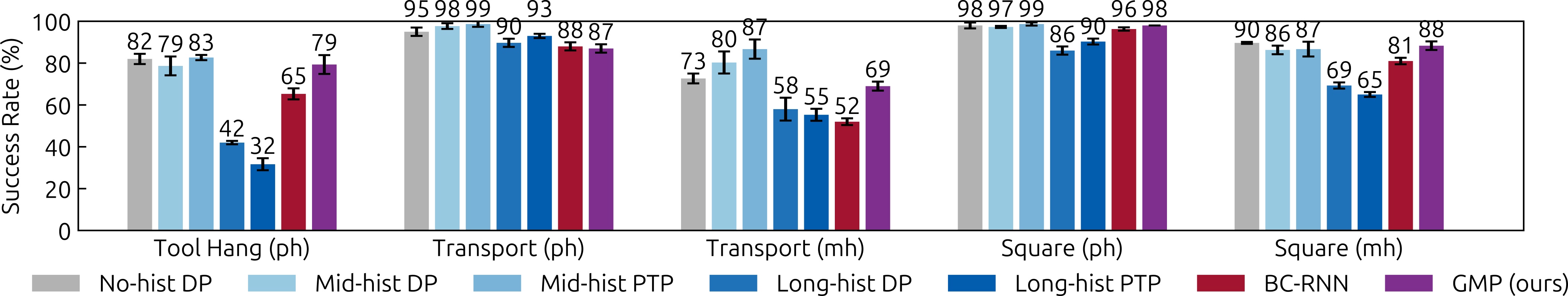}
    \end{center}
    \vspace{-4mm}
    \caption{
    \textbf{Evaluation results on RoboMimic.} We evaluate 3 tasks from the RoboMimic benchmarks~\cite{robomimic2021}: \textit{Tool Hang, Square, and Transport}. (ph, mh) indicate that the data were collected from proficient-human (ph) or multi-human (mh). While most long-history policies experience performance drops on these Markovian tasks, GMP maintains competitive performance by leveraging the gating mechanism.}
    \label{fig:exp7_robomimic}
\end{figure*}

%% file: figures/exp10_mikasa.tex
\begin{figure*}
    \begin{center}
    \includegraphics[width=0.88\textwidth]{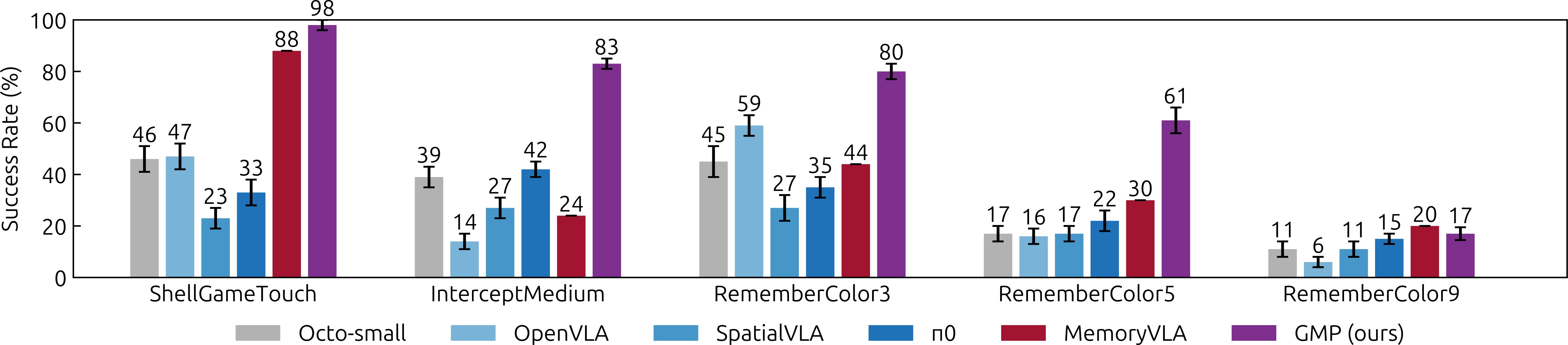}
    \end{center}
    \vspace{-4mm}
    \caption{
    \textbf{Evaluation results on MIKASA-Robo.} We evaluate 5 tasks on the MIKASA-Robo benchmarks~\cite{cherepanov2025mikasa}, out-performing prior work MemoryVLA~\cite{shi2025memoryvla} by 26.6\% on average. The baseline performance are reported in \cite{cherepanov2025mikasa} and \cite{shi2025memoryvla}.}
    \vspace{-5mm}
    \label{fig:exp10_mikasa}
\end{figure*}

%% file: figures/ablation_gate_regularization.tex
\begin{wrapfigure}{r}{0.5\linewidth}
    \centering
    \vspace{-5mm}
    \includegraphics[width=0.5\textwidth]{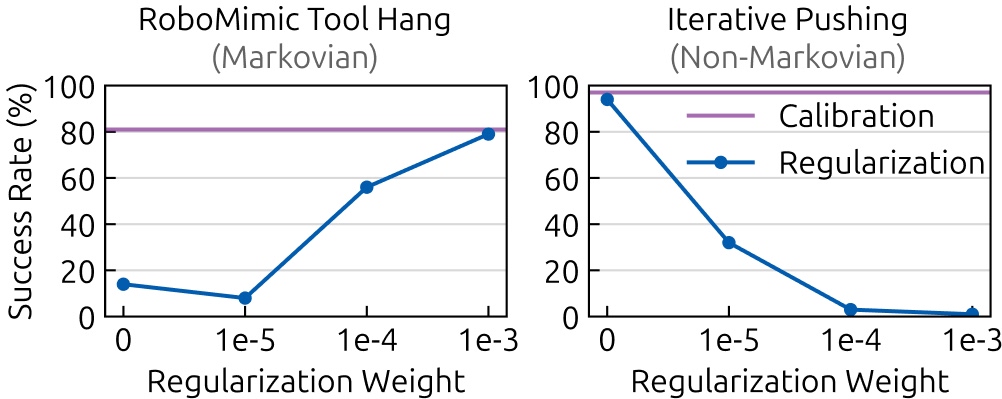}
    \caption{\textbf{Calibration of Binary Memory Gate}.}
    \label{fig:ablation_gate_regularization}
    \vspace{-5mm}
\end{wrapfigure}

%% file: figures/ablation_runtime.tex
\begin{wrapfigure}{r}{0.4\linewidth}
    \vspace{-4mm}
    \centering
    \includegraphics[width=0.37\textwidth]{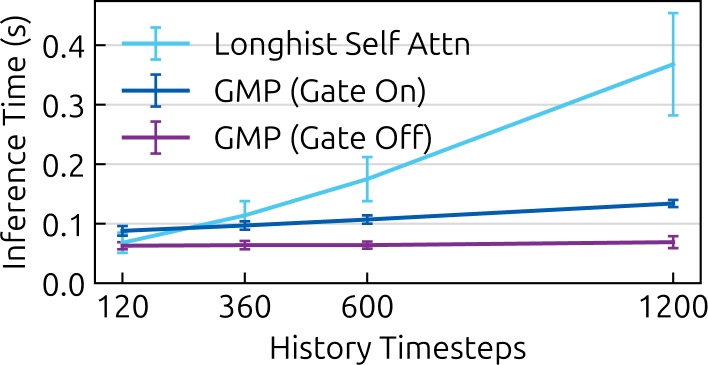}
    \vspace{-1mm}
    \caption{\textbf{Inference Time Comparison}.}
    \label{fig:inference_time_comparison}
    \vspace{-5mm}
\end{wrapfigure}

%% file: figures/ablation_noise.tex

\begin{wrapfigure}{r}{0.4\linewidth}
    \centering
    \vspace{-6mm}
    \includegraphics[width=0.35\textwidth]{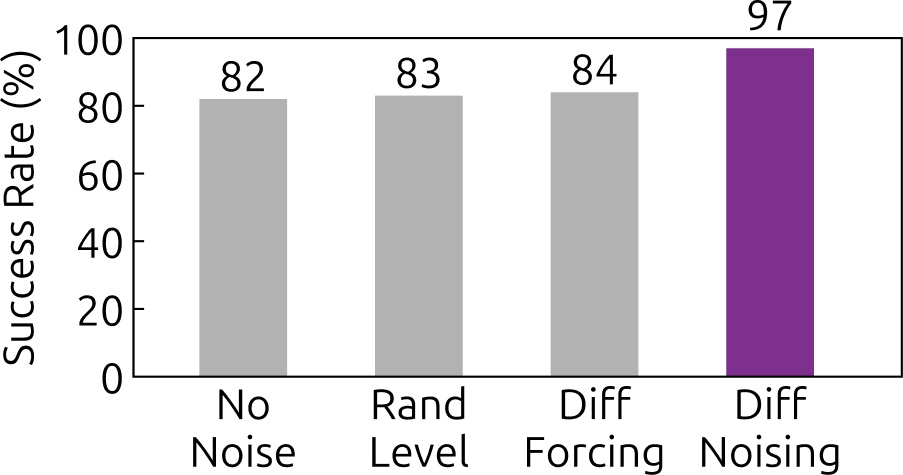}
    \vspace{-2mm}
    \caption{\textbf{Noise Injection Ablation}.}
    \label{fig:ablation_noise}
    \vspace{-4mm}
\end{wrapfigure}

%% file: text/5_final_sections.tex
\section{Limitations and Future Work}
While we show that \shortname~can effectively leverage history information through the cross-attention mechanism, the current implementation is still limited to a finite attention window. Consequently, the model cannot leverage an ``infinite'' memory context and hence may lose access to critical information over extended sequences. Future work could investigate selective caching and token replacement strategies based on importance. By dynamically managing the history buffer, the model can attend to high-value tokens from the distant past even within a fixed attention budget.

\section{Conclusion}
\vspace{-3mm}
We present Gated Memory Policy, a visuomotor policy that learns to selectively recall memory on demand. By introducing a cross-attention module for history tokens, calibrating a binary memory gate, and injecting diffusion noise into history actions, \shortname~can retrieve task-relevant information, perform in-context memorization and adaptation, remain robust to long history lengths, and maintain low computational cost. Our method outperforms other long-history baselines on the non-Markovian task suite MemMimic in both simulation and real-world settings, while avoiding performance degradation on Markovian tasks.

\section{Acknowledgments}
We thank Austin Patel for the great iPhUMI app~\cite{patel2026bpp} that enables in-the-wild data collection and deployment. We thank Chiling Han for the assistance with the baseline experiment. We thank Mengda Xu and Huy Ha for MuJoCo simulation instructions. We thank all REALab members for their valuable suggestions on manuscript writing and presentations. We also thank Yuejiang Liu, Moo Jin Kim, John Yao, Shang Yang, Genghan Zhang, and Dongchen Han for valuable discussions and feedback.

This work was supported in part by the  NSF Award \#2143601, \#2037101, and \#2132519. The views and conclusions contained herein are those of the authors and should not be interpreted as necessarily representing the official policies, either expressed or implied, of the sponsors.
We would like to thank TRI for providing the UR5 robot hardware and ARX for the X5 robot hardware. We thank Apple for providing the iPhone 15 Pro. We also thank Stanford Marlowe~\cite{kapfer2025marlowe} for providing computational resources.

%% file: text/supp.tex
%






\section{Supplementary Materials}

\subsection{How Long History Can \shortname~Recall?}
Cross-attention is scalable and efficient for long history lengths. Although the context window is not infinite, cross-attention is able to precisely attend to the most relevant observations and actions without additional supervision. In task \textbf{T1} \texttt{Match Color}, after the initial color disappears, the environment randomly waits 5--600 seconds (50--6000 frames at 10\,fps) before the final color appears. To save training GPU memory, we follow \cite{torne2025ldp} and freeze the vision encoder from the original \texttt{Match Color} task and cache all of the image features in the training dataset. After training on 500 episodes, we achieve a \textbf{99.0\%$\pm$1.0\%} success rate with a memory buffer of \textbf{6000 image frames and 6000 actions}, and the policy inference only takes \textbf{0.16s} for 8 denoising steps on an NVIDIA RTX 5090 GPU. We believe this history length suffices for a visuomotor policy to complete short-term memory tasks. For long-term memory, a higher-level VLM can handle this more efficiently via text-based representations. 

\subsection{Overlapped Trajectory Training}

Training a long-history policy jointly with a vision encoder introduces significant GPU memory overhead. Prior work~\cite{torne2025ldp} addresses this issue by first training the vision encoder with a short-history policy, then freezing the encoder while training the long-history policy. However, this prevents the vision encoder from adapting to long-term dependencies.

Inspired by the parallel training of autoregressive causal transformers \cite{vaswani2017attention}, we address this challenge by introducing \emph{Overlapped Trajectory Training}. Let an episode be a sequence of observations and actions $\mathcal{E}=\{(I_1,A_1),(I_2,A_2),\dots,(I_T,A_T)\}.$
Instead of sampling trajectories independently, we construct a trajectory batch as a collection of overlapped subsequences of \emph{history length} \(H\):
\[
\tau_s=\{(I_s,A_s),\dots,(I_{s+H-1},A_{s+H-1})\}
\] where \(s=1,\dots,T-H+1\).
Consecutive subsequences share most of their observations; e.g., \(\tau_s\) and \(\tau_{s+1}\) overlap in \(H-1\) steps. This allows each image \(I_t\) to be encoded by the vision encoder only once per forward pass through a mini-batch, with the resulting feature reused across all subsequences in the batch that include \(I_t\). After each optimization step, the encoder is updated, and in the next batch all images are re-encoded in the same manner.

As a result, memory scales with the number of \emph{unique} images per batch rather than the total number of image occurrences across \(H\)-length subsequences, yielding substantial efficiency gains while still training the vision encoder end-to-end on long-history dependencies.








\subsection{\benchname~Details}

\mypara{\benchname~In-trial Memory Tasks:} Evaluate the policy's in-context memorization -- the ability to retrieve relevant history within a single execution attempt. 
\begin{itemize}[leftmargin=12pt]
    \item T1: \texttt{Match Color (Sim)}. Shown in Fig.~\ref{fig:experiment1_match_color}, the robot picks up a cube from one of 4 bins with randomly assigned colors. Once the cube is picked up, the bin colors are shuffled, and the robot needs to put the cube back in the bin with the original color. This task tests the policy's \textbf{visual memory}. 
    \item T2: \texttt{Discrete Place Back (Sim)}. Shown in Fig.~\ref{fig:experiment2_discrete_place_back}, a cube is randomly placed in one of four bins, and the robot picks it up, holds it in the air for 2 seconds before returning it to the original bin. This task tests the policy's \textbf{spatial memory}. 
    \item T3': \texttt{Continuous Place Back (Real)}. Shown in Fig.~\ref{fig:experiment3_continuous_place_back}, a cup and a saucer are randomly placed on the table. The robot must first place the cup on the saucer and then return it to within 5 cm of its original position. This task tests the policy's \textbf{spatial memory} and \textbf{robustness to real-world noise}. 
    \item T3: In-the-wild \texttt{Flip and Place Back (Real)}. Shown in Fig.~\ref{fig:experiment3p_in_the_wild_flip_and_place_back}, after the robot picks up the cup and places it on the saucer, a human flips it over by 90 degrees. The robot must flip it back and then return it to the original position. This task tests the policy's \textbf{spatial memory} and \textbf{generalizability across unseen environments and diverse objects}.
\end{itemize}
\mypara{\benchname~Cross-trial Memory Tasks:} Evaluate whether the policy can perform in-context adaptation -- learn from past interactions and refine its actions through a trial-and-error process. Each episode fixes the object's property and allows the policy to attempt $N$ trials. An episode is considered successful if the last $M$ trials ($M<N$) are correct.
\begin{itemize}[leftmargin=10pt]
\item T4: \texttt{Iterative Pushing (Sim)}. Shown in Fig.~\ref{fig:experiment4_iterative_pushing}, in each episode, the friction coefficient of the cube is randomly sampled from 0.005 to 0.015 and is unknown during testing. The policy is given 6 trials to push the object to the target region (red box). An episode is considered successful if the cube stops within the target region in all of the last 3 trials.
\item T5: \texttt{Iterative Flinging (Sim)}. Shown in Fig.~\ref{fig:experiment5_iterative_flinging},
the robot must fling the cloth so that its far edge lands in the target (black) area. In each episode, the cloth's mass is randomly sampled between 0.1 kg and 2.0 kg, and the robot is allowed five flings. Flinging too slowly prevents the cloth from fully extending, while flinging too fast causes it to fold back on itself—both resulting in failure. An episode is considered successful if the last 3 trials are all successful.

\item T6: \texttt{Iterative Casting (Real)}. Shown in Fig.~\ref{fig:experiment6_iterative_casting}, the robot needs to cast the object \cite{lim2022planar} with an unknown friction coefficient so that the object stops sliding between two green lines. The robot stops at the same position in each trial and adaptively adjusts its casting speed. In each episode, the robot is allowed 3 casting attempts. The episode is considered successful if the last 2 trials are successful.

\end{itemize}

We report the task details, including robot observation and action spaces, for each task in our proposed \benchname. For all the simulation tasks, we use the MuJoCo simulator \cite{todorov2012mujoco} to set up the environment. Each checkpoint is evaluated on 100 episodes and we report the mean and standard deviation of the best checkpoint in 3 different training seeds. For in-domain real-world tasks, we use a UR5 robot and a WSG50 gripper with fin-ray fingers \cite{chi2024universal}. For the in-the-wild real-world task (T3), we use an ARX X5 with the same UMI setup. 

\mypara{T1: Match Color (Sim)}

\textit{Task Setup.} To demonstrate that our model attends to the correct moment rather than exhibiting the Attention Sink~\cite{xiao2024streamingllm} effect, we initialize the bin colors only after the robot approaches the cube. We randomly initialize the cube position and the colors of the 4 bins, yielding $4\times 4!=96$ different configurations. To reduce training time while still keeping the visual memory requirement, we fix the final color scheme after the robot lifts the cube, otherwise there will be $4\times 4! \times 4! =2304$ different configurations.

\textit{Data Collection.} We use a heuristic policy to collect training data. The robot starting pose, cube pose, relative grasping pose, and dropping pose are all randomly sampled within certain ranges, while the pausing pose remains fixed to avoid information leakage when training the no-memory policy. The heuristic policy applies linear interpolation between the key poses and pauses for 2 seconds after lifting the cube, ensuring that more than 2 seconds of memory are required. We use different ranges of starting and ending poses, so that even when the memory gate is off, the policy can learn whether the task is completed. We collect 300 training episodes for this task.

\textit{Training and Deployment.} The policy uses a third-person camera view as visual observation and the absolute end-effector pose for proprioceptive feedback and action commands. The policy runs at 10Hz with a 500Hz MuJoCo physics update frequency. Applying action command interpolation from 10Hz to 500Hz significantly improves simulation stability compared to directly sending pose targets at 10Hz. The action prediction horizon of all policies is 16 steps (corresponding to 1.6s), and we only execute the first 8 steps (action execution horizon) before the next policy inference.

\textit{Baseline Failure Modes.} At the pausing stage, [No-hist DP] has no memory of how long it has already paused, and thus consistently predicts a pausing trajectory within the first 8 actions. Therefore, it becomes stuck while holding the cube in the air and reaches the time limit. [Mid-hist DP] can escape the pausing trap, but still lacks sufficient memory to recall which colored bin the cube was originally on. Consequently, [Mid-hist DP] places the cube back into a random bin. [Long-hist DP] has sufficient memory to complete the task, achieving a 100\% success rate.

\mypara{T2: Discrete Place Back (Sim)}

The task description is similar to that of T1, but the bins are not initialized with different colors, thus there are only 4 different configurations. The data collection procedure, training configuration, and failure modes are identical to those of T1.

\mypara{T3': Continuous Place Back (Real)}

\input{figures/exp3_continuous_place_back.tex}

\textit{Data Collection.} We use an iPhone-based UMI gripper (iPhUMI) \cite{patel2026bpp} to manually collect 262 training episodes. We randomly place the saucer on the table, and place the cup next to the saucer in a variety of directions. For the same saucer position, we ensure the cup handle is pointing in the same direction, thus the policy cannot simply overfit to the cup handle direction and ignore the long-term dependencies. See Fig.~\ref{fig:experiment3_continuous_place_back} (b) for an example of the initial states.

\textit{Training and Deployment.} The policy takes the iPhone ultra-wide camera as input and predicts relative end-effector poses as described in the UMI paper~\cite{chi2024universal}. The action prediction frequency is 20Hz during training, and we apply a random trajectory interval (instead of a fixed 8-step interval, we sample the next trajectory randomly between 4 and 8 steps after the previous trajectory), improving the policy's temporal robustness against imperfect policy execution. During \shortname~deployment, we reduce both the history and future action frequencies to 15Hz for better stability. We predict 16 future actions and execute the first 8 before the next policy inference. We enable asynchronous policy inference (the robot keeps moving the remaining steps during policy inference) for \shortname~to improve execution speed. We apply \textit{dynamic latency matching} for a smooth transition between the executing trajectory and the new predicted trajectory. We evaluate the policy on 40 different initial states, shown in Fig.~\ref{fig:experiment3_continuous_place_back} (b) for each policy.

\textit{Baseline Failure Modes.} For the baselines [No-hist DP] and [Long-hist DP], despite slowing down the execution frequency even further to 10Hz, the robot triggers the UR5 emergency stop frequently when pressing against the table, leading to failed episodes. Moreover, for [Long-hist DP], the history proprioceptions readily go out of distribution, leading to a 0\% success rate, thus we only keep the history images as input. Additionally, [Long-hist DP] is unable to achieve asynchronous execution, as it jitters severely when the history is long. We must freeze the robot during policy inference. As a result, \shortname~achieves both higher performance and faster execution compared to the baselines.

\mypara{T3: In-the-wild Flip and Place Back (Real)}

\input{figures/supp_in_the_wild_data.tex}

\textit{Data Collection.} The dataset consists of: 1274 trajectories for only picking up and placing back the cup; 588 trajectories for only flipping the cup; 304 trajectories for both picking up and flipping the cup. We used 12 cups during data collection across more than 30 diverse scenes. Cups and environments during data collection are shown in Fig.~\ref{fig:supp_in_the_wild_data}.

\textit{Training and Deployment.} We realize that the iPhone ultra-wide camera does not provide clear visual information of whether the cup is flipped or not. Therefore, we use both the ultra-wide and the main camera for policy inference, both resized and padded to 256x256 resolution. We notice that even without gate training, the policy performs well, thus we treat it as a special case of \shortname~when gate is always on. We deploy the policy at 10Hz action frequency on an ARX X5 robot arm with \textit{dynamic latency matching}. We evaluate the policy in 6 challenging unseen environments and 10 cups, adding up to 20 episodes.

\textit{Failure Modes.} We observe that most of our policy's failure cases are due to the challenging unseen environments: on the bench (the top left image in Fig.~\ref{fig:experiment3p_in_the_wild_flip_and_place_back} (c)), the cup is very likely to fall into the gap on the bench; on the fountain (the top right image in Fig.~\ref{fig:experiment3p_in_the_wild_flip_and_place_back} (c)), the policy tends to estimate the surface height incorrectly, dropping the cup too early when putting it back. In other environments, the policy is able to complete the task with a high success rate. The baseline [Long-hist DP] overfits much more easily and fails to put the cup back to the original position after the flipping perturbation.

\mypara{T4: Iterative Pushing (Sim)}

\textit{Data Collection.} We use a heuristic policy to collect 500 training episodes. In each trial, the heuristic policy slowly approaches the cube, linearly accelerates to a pushing velocity $v$, maintains it for 0.2s, then starts to decelerate from a fixed position. After the cube stops sliding, the heuristic policy moves the robot back to the starting position, and the environment teleports the cube back to its starting position. One episode ends after the heuristic policy demonstrates a complete trial-and-error process for 6 trials.

The friction coefficient of the object is randomly sampled from 0.005 to 0.015. We empirically determine the optimal pushing velocity $v_\text{opt}$, ranging from 0.247m/s to 0.435m/s as the friction coefficient varies. The heuristic policy has access to the ground-truth object friction and approaches the optimal pushing velocity $v_\text{opt}$ within 4 trials: beginning from the middle of the pushing velocity range and adjusting in a binary search manner. Once it reaches the optimal velocity, the policy repeats it in all remaining trials. This ensures that the heuristic policy successfully pushes the object into the target region for the last 3 trials. We record the robot end-effector poses and the top-down camera view for training. Both data collection and policy inference employ 10Hz action frequency and the action space is constrained to the y axis. 

\textit{Training and Deployment.} We extend the history length of \shortname~to 240 timesteps, including $A_{t-240:t}$ and $I_{t-240:8:t}$ as history conditioning. This ensures that the policy context includes 2 complete trials, achieving higher performance than with only 120 timesteps. However, the performance degrades when extending the history length for [Long-hist DP] and [Long-hist PTP], thus we report the performance with 120 timesteps for long-history baselines. Randomizing the trajectory interval does not turn out to be as helpful as in the real-world experiments, as the simulation includes significantly less noise. During testing, by referring to the last trial, \shortname~is able to move closer to the optimal pushing velocity through the binary search behavior distilled from the heuristic policy.

\textit{Baseline Failure Modes.} [No-hist DP], [Mid-hist DP] and [Mid-hist PTP] lack sufficient memory to cover the last trial, and consequently push the object at a random velocity, unable to achieve continuous success in the last 3 trials. While [Long-hist DP] and [Long-hist PTP] are able to reach the optimal pushing velocity and succeed in a few trials, they struggle to maintain the same pushing velocity for multiple trials.

\mypara{T5: Iterative Flinging (Sim)}

\textit{Data Collection.} We also use a heuristic policy to collect 500 training episodes. In MuJoCo, we model the cloth as a $13\times13$ grid and the mass is randomly sampled from the following 7 values: 0.1kg, 0.15kg, 0.2kg, 0.3kg, 0.5kg, 1.0kg, 2.0kg, with a small random perturbation $\mathcal{N}(0, 0.001)$kg to increase diversity. In the flinging process, the heuristic policy lifts the cloth, flings it along a cubic spline trajectory with a series of waypoints, and returns to the starting position after the cloth falls on the table. To demonstrate the trial-and-error process for a cloth with unknown mass, it iteratively adjusts the time interval between the waypoints and achieves continuous success for the last 3 trials. We record the end-effector poses for both robot arms and the top-down camera view for training. Both data collection and policy inference employ a 10Hz action frequency, and the action space is constrained to the x and z axes; the cloth is welded to the robots' grippers for stability.

\textit{Training and Deployment.} To accelerate the simulation, we use the Projected Gauss-Seidel (PGS) solver in MuJoCo \cite{tassa2012pgs,todorov2012mujoco} to solve cloth dynamics. Due to numerical instability, we observe that the cloth may still perform unrealistic motions, such as penetrating into the table or entangling with itself. Therefore, we relax the success criterion to require at least 10 (out of 13) edge grid points touching the black area. Despite the simulation being imperfect, \shortname~still learns to iteratively refine its actions based on past experiences, achieving higher performance than the other long-context baselines.

\textit{Baseline Failure Modes.} As with T4, the flinging motion has higher velocity tolerance compared to pushing. [No-hist DP], [Mid-hist DP] and [Mid-hist PTP] learn to fling the cloth at a medium velocity, and are thus able to achieve success on cloths with medium mass (0.2kg, 0.3kg, 0.5kg), achieving a 54\% success rate, but fail on cloths with extremely light or extremely heavy masses.

\mypara{T6: Iterative Casting (Real)}

\textit{Data Collection.} We use a heuristic policy to collect 80 training episodes. In each episode, we randomly choose the contact surface to be Teflon fiber (low friction, optimal casting velocity 1.4m/s) or electrical tape (high friction, optimal casting velocity 2.2m/s). In each trial, the heuristic policy casts the object, stops at the same position, lifts the object after the object stops sliding, and resets to the starting position. In each episode, the heuristic policy executes 3 trials. In the first trial, the heuristic policy casts at a medium velocity (1.8m/s), leading to either overshooting (Teflon fiber) or undershooting (electrical tape). In the second and third trials, it casts with the optimal velocity of the chosen contact surface to ensure success. We use a top-down camera to record videos at 30Hz and randomly perturb the camera pose every 20 episodes to reduce overfitting. We constrain the robot movement to the y-z plane during both data collection and policy inference. We visualize the position control for different casting velocities in Fig.~\ref{fig:supp_casting_pos_control}.

\input{figures/supp_casting_pos_control.tex}

\textit{Training and Deployment.} We downsample the training data to 15Hz and execute \shortname{} and all other baselines at 15Hz for consistent dynamics. The policy predicts 25 future actions and we execute the first 15 before the next policy inference. During training, we randomize the trajectory interval from 12 to 18 steps and apply a random image latency from -0.3s to 0.3s to improve temporal robustness against the noise and delay in the real-world system. We evaluate the policy on 2 different contact surfaces and 10 episodes per surface.

\textit{Baseline Failure Modes.} We observe that [No-hist DP] consistently casts the object at the medium velocity (1.8m/s), leading to failure in 95\% of the episodes. Although Diffusion Policy is able to learn a multi-modal action distribution, the training error generally remains the lowest if it always chooses the medium velocity, and thus it collapses to a single mode. [Long-hist DP] still fails to include past proprioceptions due to real-world noise, leading to severe jittering with a 0\% success rate; therefore, we only keep history images as input. We notice that [Long-hist DP] learns to cast at the medium velocity for the first trial, and chooses a different velocity in the second and third trials. However, when switching from one trajectory (start of casting) to another (end of casting), [Long-hist DP] often gives inconsistent velocity predictions. Moreover, [Long-hist DP] fails to maintain the same casting velocity for the last two trials, resulting in a 20\% success rate.

\subsection{RoboMimic Evaluation Details}

We evaluate three tasks from RoboMimic~\cite{robomimic2021}: \texttt{Square}, \texttt{Transport}, and \texttt{Tool Hang}. 
Policies are trained for 400 epochs, with checkpoints saved every 20 epochs. Each checkpoint is evaluated on 100 episodes and we report the best success rate.

\subsection{MIKASA-Robo Evaluation Details}

\label{supp:mikasa_details}

We evaluate our method on the MIKASA-Robo benchmark~\cite{cherepanov2025mikasa} for 5 tasks: \texttt{ShellGameTouch}, \texttt{InterceptMedium}, \texttt{RememberColor3}, \texttt{RememberColor5}, and \texttt{RememberColor9}. To match the training protocol in prior work, we use 250 training episodes per task generated by an oracle state-based PPO policy with image resolution 128$\times$128. Policies are trained for 500 epochs, with checkpoints saved every 20 epochs. Each checkpoint is evaluated on 100 episodes and we report the best success rate.

We notice that the action generated by the PPO oracle policy is very noisy. Instead of using delta joint actions, we calculate the equivalent absolute joint actions and use them as training supervision. Due to the different task settings, we try different action prediction and execution horizons and report the best performing checkpoint. To match the training protocol in prior work, we only use 250 episodes per task, which is insufficient for the remember color tasks, especially for \texttt{RememberColor9}. More training episodes (e.g., 1000) will lead to significantly better performance. Another approach to achieve higher performance is to use rule-based heuristics or human teleoperation to generate training data, which has much lower action noise.

\subsection{Baseline Details}

\begin{itemize}[leftmargin=12pt]
    \item \textbf{No-hist DP}: A standard diffusion policy with a Transformer backbone that includes only the most recent image $I_t$ and robot proprioception $P_t$ as input. 

    \item \textbf{Mid-hist DP}: Extend DP history input length from 1 to 16 steps: using proprioception $P_{t-16:t}$ and images $I_{t-16:t}$ as input.
    
    \item \textbf{Mid-hist PTP \cite{torne2025ldp}}:  a recent method enabling long-history policy inference. It predicts 16 past actions $A_{t-16:t}$ together with 16 future actions $A_{t:t+16}$. This configuration matches the longest context length in the original PTP implementation.

    \item \textbf{Long-hist DP}: Further extend the DP history length to 120. Due to the GPU memory limit, we sample every 8th image: using proprioception $P_{t-120:t}$ and images $I_{t-120:8:t}$ as input.

    \item \textbf{Long-hist PTP}: Based on [Long-hist DP], it predicts 120 past actions $A_{t-120:t}$ together with 16 future actions $A_{t:t+16}$. This history length is 7$\times$ longer than the length reported in \cite{torne2025ldp}.

    \item \textbf{BC-RNN \cite{robomimic2021}}: An LSTM-based RNN policy, predicting an 8-step action chunk $A_{t:t+8}$ at every inference step. We do not reset the RNN states, thus the history length is unlimited.
    
    \item \textbf{\shortname~(Ours)}: We match the history configuration in [Long-hist DP], using 120 history actions $A_{t-120:t}$ and the corresponding 15 images $I_{t-120:8:t}$ as the cross-attention input. 
    
\end{itemize}

For iterative tasks \textbf{T4-6}, we further extend the history length of GMP and the long-context baselines.

\subsection{Visualization of the Memory Gate Calibration}

In this section, we visualize the statistics of the memory gate calibration process for one example episode per task. This demonstrates that the calibration process can identify whether a state requires memory and generate corresponding labels, ensuring the functionality of the memory gate during deployment.

After training two policies with and without memory, $\blue{\pi_{\mathrm{mem}}}$ and $\gray{\pi}$, we sample 500 trajectory batches from each episode for $\blue{\pi_{\mathrm{mem}}}$ and 20000 samples from each episode for $\gray{\pi}$ on the validation set to calculate the average action prediction error. To reduce noise from outliers, we apply a sliding-window average of 20 timesteps to both $\blue{\delta^\mathrm{mem}_t}$ and $\gray{\delta_t}$; Fig.~\ref{fig:supp_gate_label_statistics} shows the raw errors and labels before smoothing, and the smoothing widens the active-label regions. Finally, we label $\mu_t$ as 1 if $\gray{\delta_t} > \theta \blue{\delta^\mathrm{mem}_t}$ for a threshold $\theta=10.0$ across all tasks, otherwise $\mu_t=0$. We visualize the action prediction errors and memory gate labels along with task progress in Fig.~\ref{fig:supp_gate_label_statistics}.

\input{figures/supp_gate_label_statistics.tex}

We observe that the memory gate is on (i.e., $\mu_t=1$) earlier than the moments that require memory. This is because we are calculating the error for the entire predicted action trajectory $\delta_t=\Vert\hat{A}_{t:t+h} - A_{t:t+h}\Vert_2$, which aligns with the policy deployment requirement---the policy should recall memory when predicting actions right before the critical moment.
For the RoboMimic tasks, we observe that the prediction error with or without memory is similar, indicating that the memory gate can identify the critical moments that require memory and omit memory recall when it is not needed.

\subsection{Why Not Use Continuous Gating}

While prior methods mostly use a continuous gate value to control the amount of history to be recalled, we find that simply training a continuous gate together with the policy model is insufficient for memory usage control. Similar to \textbf{Finding 4}, without applying a regularization loss term, the gate value tends to converge to 1, leading to poor performance on Markovian tasks, as shown in Fig.~\ref{fig:supp_continuous_gate_regularization}. The success rate difference in \texttt{Tool Hang} is not as significant as that of the binary gate, because we do not need to apply Straight-Through Estimator (STE) during training and the training gradients are more stable. However, when applying regularization, there is also a significant trade-off between performance on Markovian and non-Markovian tasks, making it difficult to find a regularization weight that works well across all tasks.

\input{figures/supp_continuous_gate_regularization.tex}

While it is possible to apply the same calibration process to a continuous gate, replacing the BCE loss with an MSE loss and keeping all other stages the same, we find that the continuous-gated policy works similarly to the binary-gated policy. However, using continuous values limits the ability to skip memory recall when unnecessary, thereby preventing speedup of policy inference.

\subsection{Training Cost and Calibration Overhead}

Full model convergence is not required for label generation. Therefore, a checkpoint from the first few epochs is sufficient to observe prediction error differences. We also don't need to run the calibration process on the full dataset, as a subset of the validation set is sufficient. We report the GPU-hours on H100 GPUs required to reach the best success rate on the most challenging task, \texttt{Iterative Pushing}.

\begin{center}
{\small
\begin{tabular}{ccccc}
\hline
\textbf{Train $\pi$} & \textbf{Train $\pi_\mathrm{mem}$} & \textbf{Rollout} & \textbf{Train Gate} & \textbf{Train GMP} \\
\hline
1.2h & 6.7h & 6.4h & 1.3h & 31.2h \\
\hline
\end{tabular}
}
\end{center}

However, we do notice that the gate is not necessary for all tasks. In tasks that do not require high-precision actions, for example \texttt{MIKASA-Robo} tasks and \texttt{Continuous Place Back}, the policy without the gate already achieves decent performance. A general rule of thumb is to first train the policy without the gate and run evaluation. If there are clear signals that the policy fails because of overfitting, for example in the \texttt{RoboMimic Tool Hang} task, then we train the gate and consequently the gated policy.

%% file: figures/exp3_continuous_place_back.tex
\begin{figure*}
    \begin{center}
    \includegraphics[width=\textwidth]{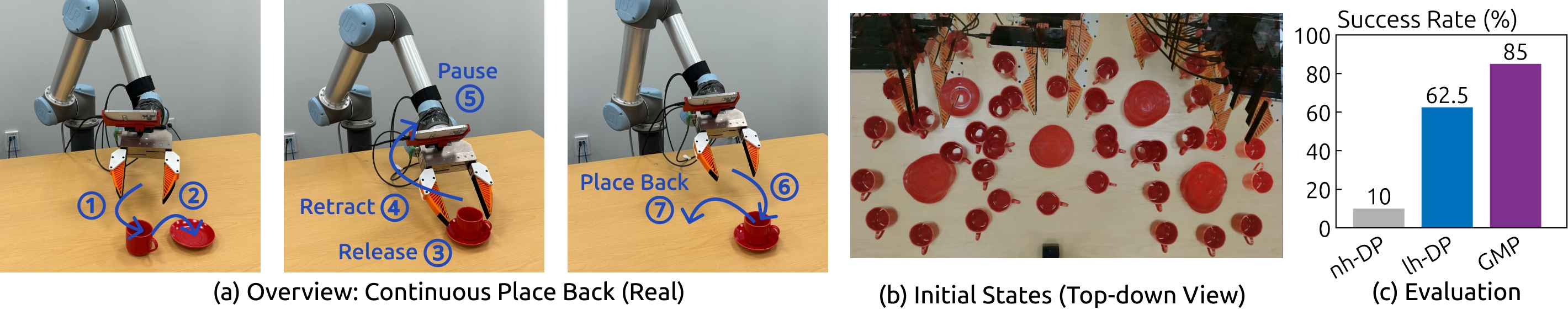}
    \end{center}
    \vspace{-3mm}
    \caption{\textbf{Task 3': Continuous Place Back (Real)}. 
    \textbf{(a)} The robot picks up the cup, places it on the saucer, picks it up again, and returns it to the original position. 
    \textbf{(b)} Initial positions of the cup and saucer tested across the continuous workspace. 
    }
    \vspace{-2mm}
    \label{fig:experiment3_continuous_place_back}
\end{figure*}

%% file: figures/supp_in_the_wild_data.tex
\begin{figure}[ht]
    \begin{center}
    \includegraphics[width=0.7\textwidth]{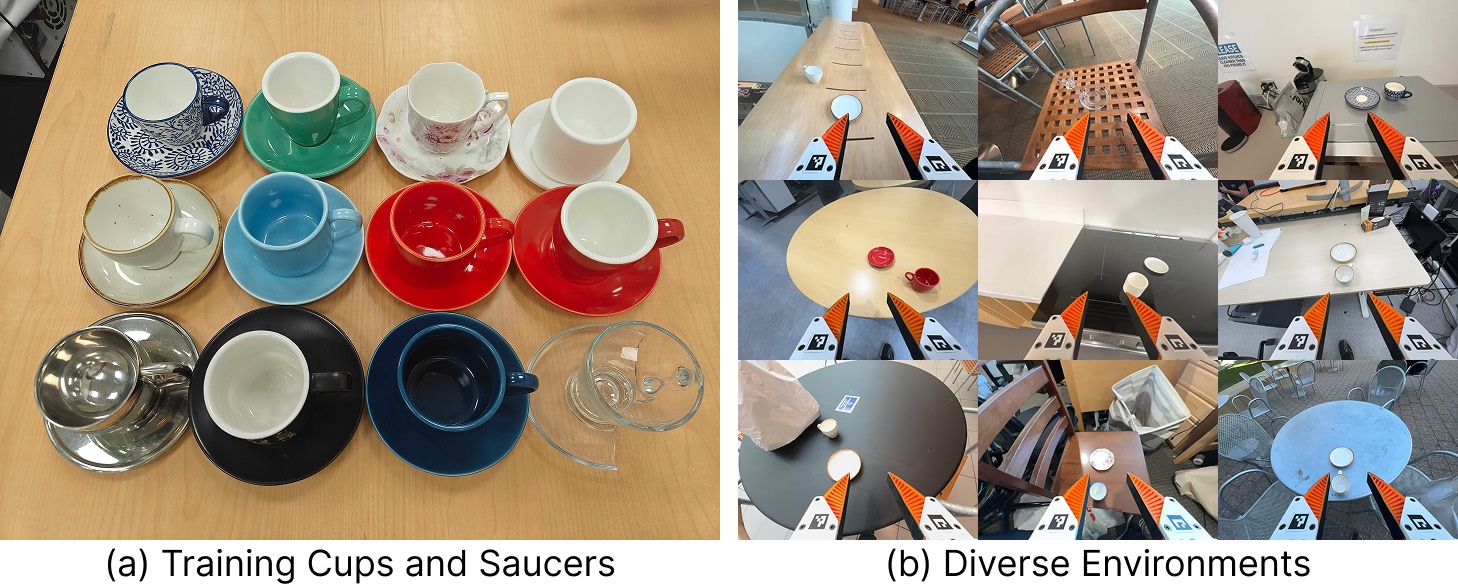}
    \end{center}
    \vspace{-5pt}
    \caption{\textbf{Training objects and scenes for In-the-wild Flip and Place Back}. We show the cups used for data collection in the real-world \texttt{Flip and Place Back} task and a subset of the environments used for data collection.}
    \label{fig:supp_in_the_wild_data}
    \vspace{-3mm}
\end{figure}

%% file: figures/supp_casting_pos_control.tex
\begin{figure}[ht]
    \begin{center}
    \includegraphics[width=0.4\textwidth]{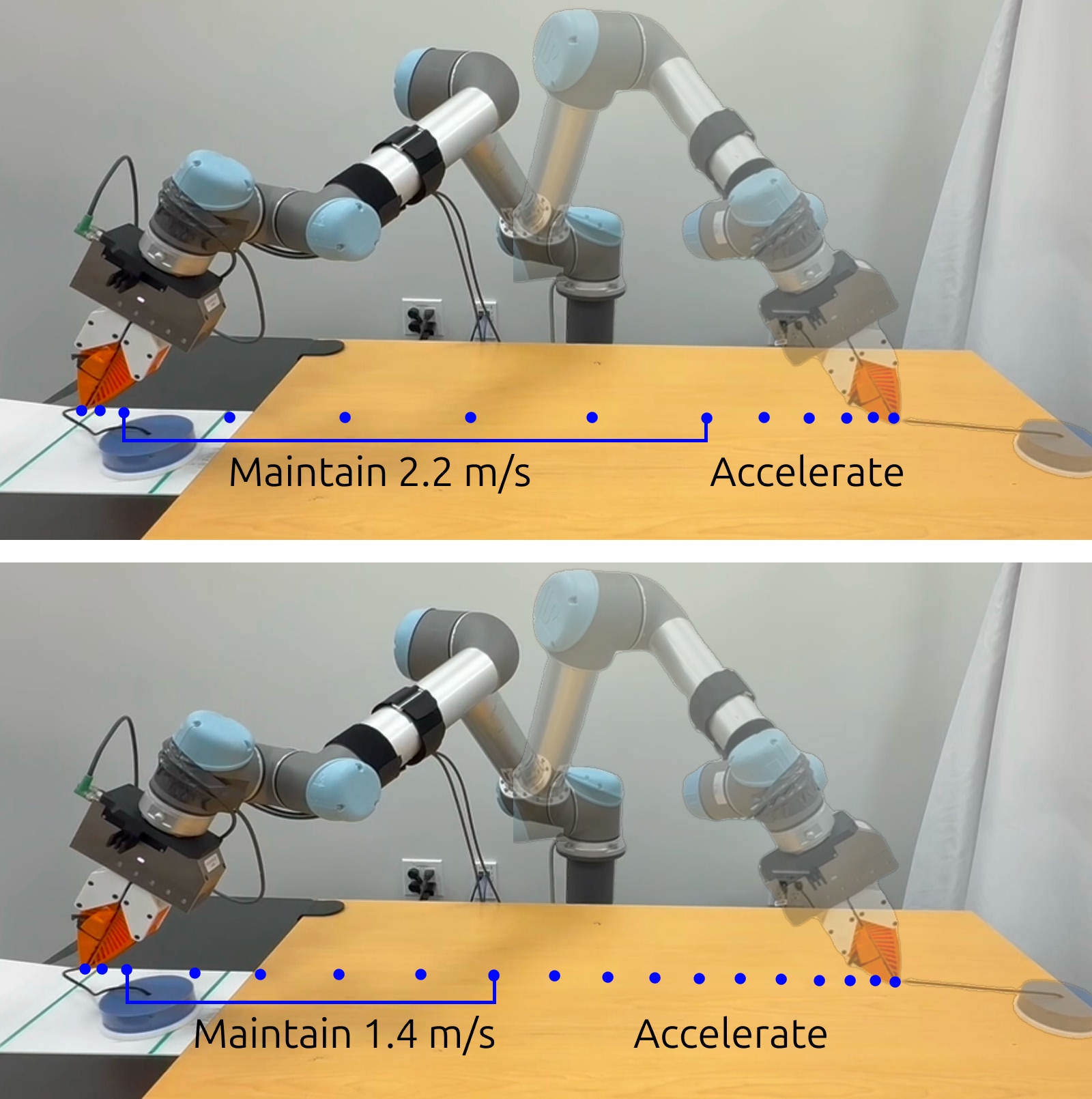}
    \end{center}
    \vspace{-5pt}
    \caption{\textbf{Position Control for Different Casting Velocities}.
    We present examples of waypoint-based position control to achieve different casting velocities. While starting and ending at the same position and running at the same frequency, the heuristic policy adjusts the waypoint distribution to modify the casting velocity. Before the robot decelerates, sparser waypoints lead to faster casting velocities, while denser waypoints lead to slower casting velocities.}
    \label{fig:supp_casting_pos_control}
    \vspace{-3mm}
\end{figure}

%% file: figures/supp_gate_label_statistics.tex
\begin{figure}[ht]
    \begin{center}
    \includegraphics[width=0.5\textwidth]{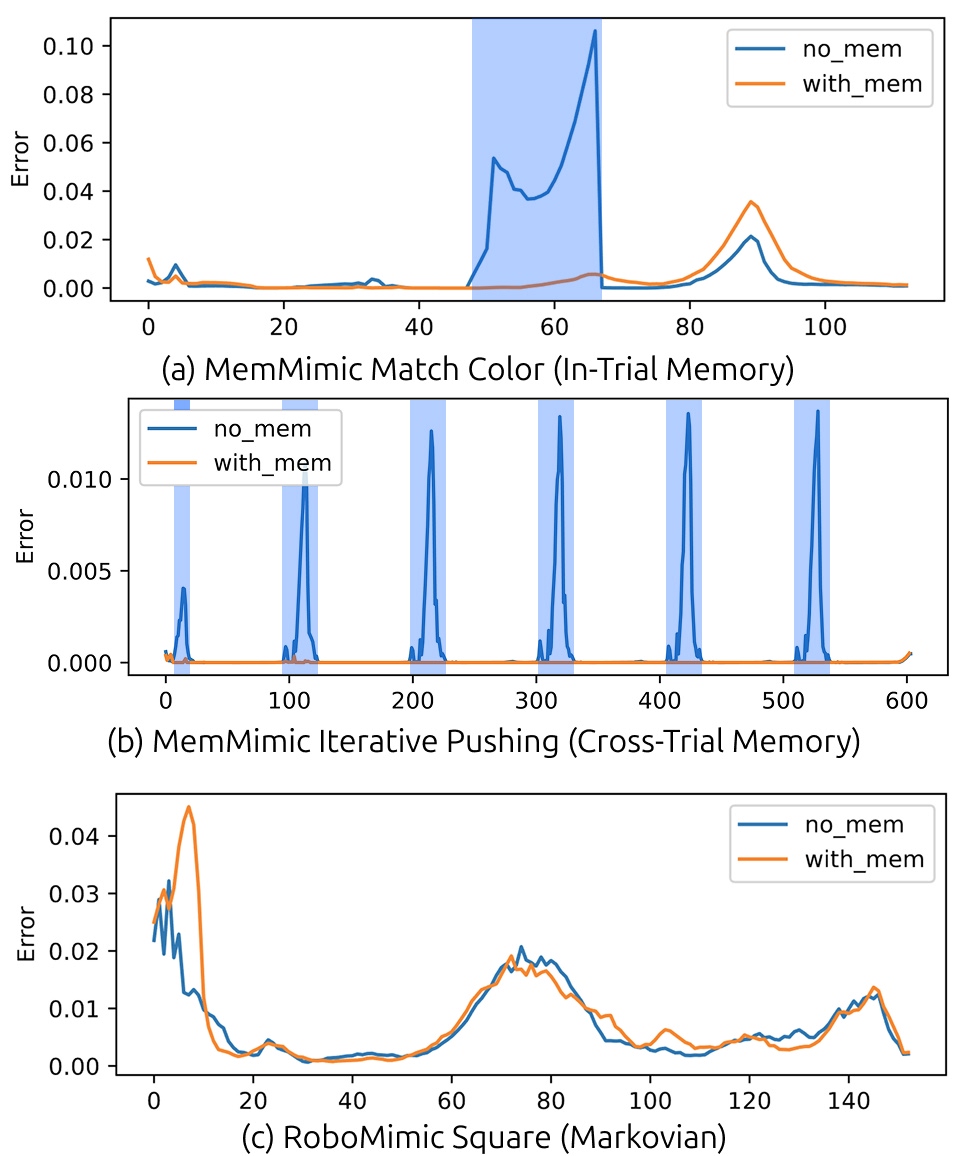}
    \end{center}
    \vspace{-5pt}
    \caption{\textbf{Example Statistics of Memory Gate Label Generation}. We present the action prediction errors for both the no-memory policy ($\gray{\delta_t}$) and memory policy ($\blue{\delta^\mathrm{mem}_t}$) for one example episode per memory requirement category. The x-axis represents the timestep of each episode. We use a blue background to indicate when the memory gate label is set to 1, and no background when it is set to 0. We observe that in \textbf{(a)} \texttt{Match Color}, $\gray{\delta_t} > \theta \blue{\delta^\mathrm{mem}_t}$ in the middle of the episode when the robot is placing the cube back; in \textbf{(b)} \texttt{Iterative Pushing}, $\gray{\delta_t} > \theta \blue{\delta^\mathrm{mem}_t}$ in every trial before the robot pushes; while in \textbf{(c)} \texttt{Square}, the action prediction error is similar between the two policies, indicating that the memory gate should remain off throughout.}
    \label{fig:supp_gate_label_statistics}
    \vspace{-3mm}
\end{figure}

%% file: figures/supp_continuous_gate_regularization.tex
\begin{figure}[ht]
    \begin{center}
    \includegraphics[width=0.5\textwidth]{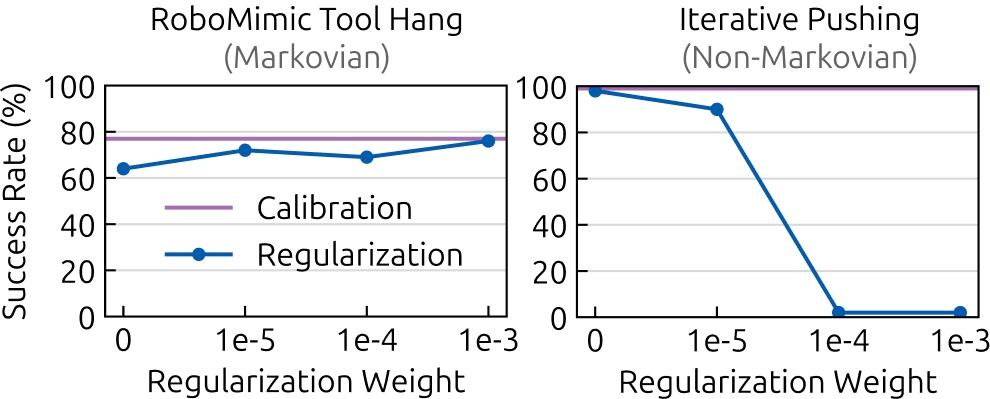}
    \end{center}
    \caption{\textbf{Continuous Memory Gate Ablation}.
    When training a continuous memory gate jointly with the policy, using no regularization, which encourages the gate to stay on, degrades performance on Markovian tasks, while using high regularization, which encourages the gate to be off, hurts performance on non-Markovian tasks.}
    \label{fig:supp_continuous_gate_regularization}
    \vspace{-3mm}
\end{figure}